\documentclass[11pt,a4paper]{article}
\usepackage[hyperref]{emnlp2020}
\usepackage{times}
\usepackage{latexsym}
\usepackage{tabularx}
\usepackage{graphicx}
\usepackage{float}
\usepackage{enumerate}
\usepackage{amssymb,amsmath}
\usepackage{color}
\usepackage{bm}
\usepackage{makecell}
\usepackage{multi row}
\usepackage{array}
\usepackage{resizegather}
\usepackage{refcount}
\usepackage{fixfoot}
\usepackage{hyperref}
\usepackage{url}
\usepackage{subcaption}  
\usepackage{bbm}

\usepackage{algorithm}
\usepackage{algorithmicx}
\usepackage{algpseudocode}

\usepackage{microtype}

\newif\ifcomment
\commenttrue
\ifcomment
\newcommand{\kc}[1]{\textcolor{red}{KC: #1}}
\newcommand{\tl}[1]{\textcolor{blue}{TL: #1}}
\newcommand{\ql}[1]{\textcolor{cyan}{QL: #1}}
\newcommand{\cm}[1]{\textcolor{orange}{CM: #1}}
\definecolor{CMpurple}{rgb}{0.6,0.18,0.64}
\newcommand\cmm[1]{\marginpar{\tiny\raggedright\textcolor{CMpurple}{\textsf{\bfseries CM\@: #1}}}}
\else
\newcommand{\kc}[1]{}
\newcommand{\tl}[1]{}
\newcommand{\ql}[1]{}
\newcommand{\cm}[1]{}
\newcommand{\cmm}[1]{}
\fi

\newcommand\tstrut{\rule{0pt}{2.6ex}}         
\newcommand\bstrut{\rule[-1.0ex]{0pt}{0pt}}   
\newcommand{\thinline}{\Xhline{1.5\arrayrulewidth}}
\newcommand{\thickline}{\Xhline{2.5\arrayrulewidth}}
\newcommand{\tsep}	{\bstrut \\ \thinline}

\newcommand{\ttop}{\thickline}
\newcommand{\tbottom}{\bstrut \\ \thickline}
\newcommand{\replace}[3] {
    \textsc{replace}(#1, #2, #3)
}

\newcommand{\alns}[1] {
	\begin{align*} #1 \end{align*}
}

\newcommand{\bfx}{\bm{x}^\text{noised}}

\newcommand{\Z}{Z_\theta(\cntxt)}

\newcommand{\bx}{\bm{x}}
\newcommand{\bw}{\bm{w}}
\newcommand{\bh}{\bm{h}}
\newcommand{\cntxt}{\bx_{\backslash t}}
\newcommand{\vocab}{\mathcal{V}}
\newcommand{\up}{\hat{p}_\theta}
\newcommand{\pdata}{p_\text{data}}
\newcommand{\E} {\mathop{\mathbb{E}}}
\newcommand{\hbx}{\hat{\bm{x}}}
\newcommand{\mask}{\texttt{[MASK]}}
\newcommand{\hf} { \overrightarrow{\bh} }
\newcommand{\hb} { \overleftarrow{\bh} }
\newcommand{\bs} {\bm{s}}
\DeclareMathOperator*{\argmax}{arg\,max}

\usepackage{etoolbox}
\newcommand{\zerodisplayskips}{
\addtolength{\abovedisplayskip}{-1pt}
\addtolength{\belowdisplayskip}{-1pt}
\addtolength{\abovedisplayshortskip}{-1pt}
\addtolength{\belowdisplayshortskip}{-1pt}
}
\appto{\normalsize}{\zerodisplayskips}
\appto{\small}{\zerodisplayskips}
\appto{\footnotesize}{\zerodisplayskips}

\aclfinalcopy

\title{Pre-Training Transformers as Energy-Based Cloze Models}

\author{Kevin Clark$^1$ \hspace{3mm} Minh-Thang Luong$^2$ \hspace{3mm} Quoc V. Le$^2$ \hspace{3mm} Christopher D. Manning$^1$\\
   $^1$Stanford University \hspace{6mm} $^2$Google Brain \\
   {\tt kevclark@cs.stanford.edu, thangluong@google.com} \\
   {\tt qvl@google.com, manning@cs.stanford.edu} \\
 }
 
\date{}

\begin{document}
\maketitle
\begin{abstract}
We introduce Electric, an energy-based cloze model for representation learning over text. 
Like BERT, it is a conditional generative model of tokens given their contexts. 
However, Electric does not use masking or output a full distribution over tokens that could occur in a context.
Instead, it assigns a scalar energy score to each input token indicating how likely it is given its context.
We train Electric using an algorithm based on noise-contrastive estimation and elucidate how this learning objective is closely related to the recently proposed ELECTRA pre-training method. 
Electric performs well when transferred to downstream tasks and is particularly effective at producing likelihood scores for text: it re-ranks speech recognition n-best lists better than language models and much faster than masked language models. Furthermore, it offers a clearer and more principled view of what ELECTRA learns during pre-training.
\end{abstract}

\section{Introduction}

The cloze task \citep{Taylor1953ClozePA} of predicting the identity of a token given its surrounding context has proven highly effective for representation learning over text. BERT \citep{devlin2018bert} implements the cloze task by replacing input tokens with [MASK], but this approach incurs drawbacks in efficiency (only 15\% of tokens are masked out at a time) and introduces a pre-train/fine-tune mismatch where BERT sees [MASK] tokens in training but not in fine-tuning. 
ELECTRA \citep{clark2020electra} uses a different pre-training task that alleviates these disadvantages. Instead of masking tokens, ELECTRA replaces some input tokens with fakes sampled from a small generator network. The pre-training task is then to distinguish the original vs. replaced tokens. 
While on the surface it appears quite different from BERT, in this paper we elucidate a close connection between ELECTRA and cloze modeling.
In particular, we develop a new way of implementing the cloze task using an energy-based model (EBM).
Then we show the resulting model, which we call Electric, is closely related to ELECTRA, as well as being useful in its own right for some applications.\footnote{Code is available at \url{https://github.com/google-research/electra}}  

EBMs learn an energy function that assigns low energy values to inputs in the data distribution and high energy values to other inputs.  
They are flexible because they do not have to compute normalized probabilities. 
For example, Electric does not use masking or an output softmax, instead producing a scalar energy score for each token where a low energy indicates the token is likely given its context.
Unlike with BERT, these likelihood scores can be computed simultaneously for all input tokens rather than only for a small masked-out subset. 
We propose a training algorithm for Electric that efficiently approximates a loss based on noise-contrastive estimation \citep{Gutmann2010NoisecontrastiveEA}. 
Then we show that this training algorithm is closely related to ELECTRA; in fact, ELECTRA can be viewed as a variant of Electric using negative sampling instead of noise-contrastive estimation.

\begin{figure*}[tb]
\begin{center}
\includegraphics[width=0.95\textwidth]{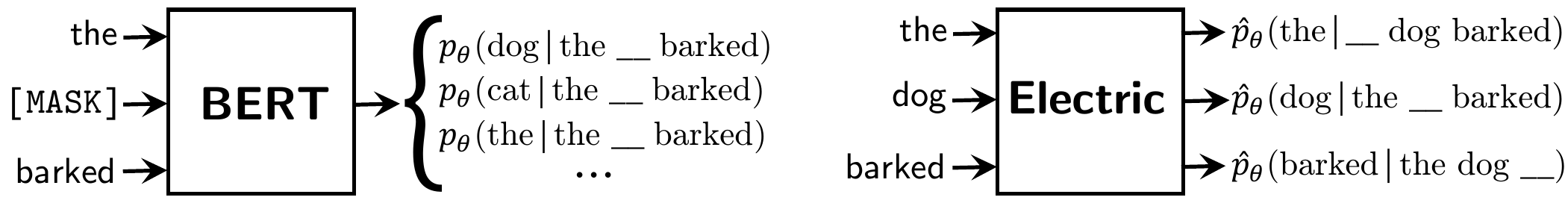}
\end{center}
\vspace{-1.5mm}
\caption{Comparison of BERT and Electric\@. Both model the probability of a token given its surrounding context, but BERT produces a full output distribution over tokens only for masked positions while Electric produces un-normalized probabilities (but no full distribution) for all input tokens.}
\vspace{-1.5mm}
\label{fig:overview}
\end{figure*}

We evaluate Electric on GLUE \citep{wang2018glue} and SQuAD \citep{Rajpurkar2016SQuAD10}, where Electric substantially outperforms BERT but slightly under-performs ELECTRA.  
However, Electric is particularly useful in its ability to efficiently produce pseudo-likelihood scores \citep{Salazar2019MaskedLM} for text: 
Electric is better at re-ranking the outputs of a speech recognition system than GPT-2 \citep{Radford2019LanguageMA} and is much faster at re-ranking than BERT because it scores all input tokens simultaneously rather than having to be run multiple times with different tokens masked out.
In total, investigating Electric leads to a more principled understanding of ELECTRA and our results suggest that EBMs are a promising alternative to the standard generative models currently used for language representation learning.

\section{Method}

BERT and related pre-training methods \cite{baevski2019cloze, liu2019roberta,  lan2019albert}
train a large neural network to perform the cloze task.
These models learn the probability $\pdata(x_t | \cntxt)$ of a token $x_t$ occurring in the surrounding context $\cntxt = [x_1, ..., x_{t- 1}, x_{t+1}, ..., x_n]$.
Typically the context is represented as 
the input sequence with $x_t$ replaced by a special \mask placeholder token. 
This masked sequence is encoded into vector representations by a transformer network \cite{Vaswani2017AttentionIA}.
Then the representation at position $t$ is passed into a softmax layer to produce a 
distribution over tokens $p_\theta(x_t | \cntxt)$ for the position.

\subsection{The Electric Model}

Electric also models $\pdata(x_t | \cntxt)$, but does not use masking or a softmax layer.
Electric first maps the unmasked input $\bx = [x_1, ..., x_n]$ into contextualized vector representations $\bh(\bx) = [\bh_1, ..., \bh_n]$ using a transformer network.
The model assigns a given position $t$ an energy score
\alns{
E(\bx)_t = \bw^T \bh(\bx)_t
}
using a learned weight vector $\bw$.
The energy function defines a distribution over the possible tokens at position $t$ as
\alns{
    p_\theta(x_t | \cntxt) &= \exp{(-E(\bx)_t)} / Z(\cntxt) \\
    &\hspace{-8mm} = \frac{\exp{(-E(\bx)_t)}}{\sum_{x' \in \vocab} \exp{(-E(\replace{\bx}{t}{x'})_t)}}
}
where $\replace{\bx}{t}{x'}$ denotes replacing the token at position $t$ with $x'$ and $\vocab$ is the vocabulary, in practice usually word pieces \cite{Sennrich2016NeuralMT}.
Unlike with BERT, which produces the probabilities for all possible tokens $x'$ using a softmax layer, a candidate $x'$ is passed in as {\it input} to the transformer. 
As a result, computing $p_\theta$ is prohibitively expensive because the partition function $\Z$ requires running the transformer $|\vocab|$ times;
unlike most EBMs, the intractability of $\Z$ is more due to the expensive scoring function rather than having a large sample space. 

\subsection{NCE Loss}
As computing the exact likelihood is intractable, training energy-based models such as Electric with standard maximum-likelihood estimation is not possible.
Instead, we use (conditional) Noise-Contrastive Estimation (NCE) \citep{Gutmann2010NoisecontrastiveEA, Ma2018NoiseCE}, which provides a way of efficiently training an un-normalized model that does not compute $\Z$.
NCE learns the parameters of a model by defining a binary classification task where samples from the data distribution have to be distinguished from samples generated by a noise distribution $q(x_t|\bx_{\backslash t})$. First, we define the un-normalized output
\vspace{-0.5mm}
\alns{
    \up(x_t | \cntxt) = \exp{(-E(\bx)_t)} 
}
\noindent Operationally, NCE can be viewed as follows: \vspace{0mm}
\begin{itemize}
    \item A positive data point is a text sequence $\bx$ from the data and position in the sequence $t$. \vspace{-1.2mm}
    \item A negative data point is the same except $x_t$, the token at position $t$, is replaced with a noise token $\hat{x}_t$ sampled from $q$. \vspace{-1.2mm}
    \item Define a binary classifier $D$ that estimates the probability of a data point being positive as\vspace{-1mm}
    \alns{\frac{n\cdot\up(x_t | \cntxt)}{n\cdot\up(x_t | \cntxt) + k\cdot q(x_t | \cntxt)}} \vspace{-5mm}
    \item The binary classifier is trained to distinguish positive vs negative data points, with $k$ negatives sampled for every $n$ positive data points. \vspace{-1.2mm}
\end{itemize}
\vspace{-1.2mm}
Formally, the NCE loss $\mathcal{L}(\theta)$ is
\alns{
n\cdot\E_{\bx, t} &\left[-\log{\frac{n \cdot \up(x_t | \cntxt)}{n \cdot \up(x_t | \cntxt) + k\cdot q(x_t | \cntxt)}} \right] + \\
k\cdot \hspace{-1.5mm} \E_{\substack{\bx, t \\ \hat{x}_t \sim q}} &\left[ -\log{\frac{k\cdot q(\hat{x}_t | \cntxt)}{n \cdot \up(\hat{x}_t | \cntxt) + k\cdot q(\hat{x}_t | \bx_{\backslash t})}} \right]
}

\noindent This loss is minimized when $\up$ matches the data distribution $p_\text{data}$ \cite{Gutmann2010NoisecontrastiveEA}.
A consequence of this property is that the model learns to be self-normalized such that $\Z = 1$. 

\subsection{Training Algorithm}
To minimize the loss, the expectations could be approximated by sampling as shown in Algorithm~\ref{alg:naive}.
\begin{algorithm}
    \caption{Naive NCE loss estimation}
    \label{alg:naive}
    \begin{algorithmic}
    \State \textbf{Given:} Input sequence $\bx$, number of negative samples $k$, noise distribution $q$, model $\up$.
    \State \textbf{1.} Initialize the loss as \phantom{aaaaaaaaaaaaaaaa} $\sum_{t=1}^n \left( -\log{\frac{n \cdot \up(x_t | \bx_{\backslash t})}{n \cdot \up(x_t | \bx_{\backslash t}) + k\cdot q(x_t | \bx_{\backslash t})}} \right)$. 
    \State \textbf{2.} Sample $k$ negative samples according to $t \sim \text{unif}\{1, n\}$, $\hat{x}_t \sim q(x_t | \bx_{\backslash t})$. 
    \State \textbf{3.} For each negative sample, add to the loss $-\log{\frac{k \cdot q(\hat{x}_t | \bx_{\backslash t})}{n \cdot \up(\hat{x}_t | \bx_{\backslash t}) + k\cdot q(\hat{x}_t | \bx_{\backslash t})}}$.
    \end{algorithmic}
    \vspace{-1mm}
\end{algorithm}
Taking the gradient of this estimated loss produces an unbiased estimate of $\nabla_\theta \mathcal{L}(\theta)$. 
However, this algorithm is computationally expensive to run, since it requires $k + 1$ forward passes through the transformer to compute the $\up$s (once for the positive samples and once for each negative sample). 
We propose a much more efficient approach that replaces $k$ input tokens with noise samples {\it simultaneously} shown in Algorithm~\ref{alg:efficient}. 
\begin{algorithm}
    \caption{Efficient NCE loss estimation}
    \label{alg:efficient}
    \begin{algorithmic}
    \State \textbf{Given:} Input sequence $\bx$, number of negative samples $k$, noise distribution $q$, model $\up$.  
    \State \textbf{1.} Pick $k$ unique random positions $R = \{r_1, ..., r_k\}$ where each $r_i$ is $1 \leq r_i \leq n$.
    \State \textbf{2.} Replace the $k$ random positions with negative samples: $\hat{x}_i \sim q(x_i | \bx_{\backslash i}) \text{ for } i \in R,$\\ $\bfx = \replace{\hbx}{R}{\hat{X}}$.
    \State \textbf{3.} For each position $t=1 \text{ to } n$: add to the loss
    $
    \begin{array}{l l}
    -\log{\frac{k \cdot q(\hat{x}_t | \bx_{\backslash t})}{(n - k)\cdot \up(\hat{x}_t | \bfx_{\backslash t}) + k\cdot q(\hat{x}_t | \bx_{\backslash t})}} & \text{if } t \in R\\
    -\log{\frac{(n - k)\cdot \up(x_t | \bfx_{\backslash t})}{(n - k)\cdot \up(x_t | \bfx_{\backslash t}) + k\cdot q(x_t | \bx_{\backslash t})}}              & \text{otherwise}
    \end{array}
    $
    \end{algorithmic}
    \vspace{-1mm}
\end{algorithm}
It requires just one pass through the transformer for $k$ noise samples and $n - k$ data samples.
However, this procedure only truly minimizes $\mathcal{L}$ if $\up(x_t | \bx_{\backslash t}) = \up(x_t | \bx^{\text{noised}}_{\backslash t})$. To apply this efficiency trick we are making the assumption they are approximately equal, which we argue is reasonable because (1) we choose a small $k$ of $\lceil0.15n\rceil$ and (2) $q$ is trained to be close to the data distribution (see below). This efficiency trick is analogous to BERT masking out multiple tokens per input sequence. 

\subsection{Noise Distribution}
The noise distribution $q$ comes from a neural network trained to match $p_\text{data}$.
NCE commonly employs this idea to ensure the classification task is sufficiently challenging for the model \citep{Gutmann2010NoisecontrastiveEA,Wang2018LearningNT}. 
In particular, we use a two-tower cloze model as proposed by \citet{baevski2019cloze}, which is more accurate than a language model because it uses context to both sides of each token.
The model runs two transformers $T_{\textsc{ltr}}$ and $T_{\textsc{rtl}}$ over the input sequence. These transformers apply causal masking so one processes the sequence left-to-right and the other operates right-to-left.
The model's predictions come from a softmax layer applied to the concatenated states of the two transformers:
\alns{
    &\hf = T_{\textsc{ltr}}(\bx), \hspace{3mm} \hb = T_{\textsc{rtl}}(\bx) \\[-2pt]
    &\vspace{-15mm}q(x_t | \cntxt) = \text{softmax}(\bm{W} [\hf_{t - 1}, \hb_{t + 1}])_{x_t} 
}
\noindent The noise distribution is trained simultaneously with Electric using standard maximum likelihood estimation over the data.
The model producing the noise distribution is much smaller than Electric to reduce the computational overhead.

\subsection{Connection to ELECTRA}
Electric is closely related to the ELECTRA pre-training method. 
ELECTRA also trains a binary classifier (the ``discriminator") to distinguish data tokens from noise tokens produced by a ``generator" network. 
However, ELECTRA's classifier is simply a sigmoid layer on top of the transformer: it models the probability of a token being negative (i.e., as replaced by a noise sample) as $\sigma(E(\bx)_t)$ where $\sigma$ denotes the sigmoid function.
Electric on the other hand models this probability as 
\alns{
&\frac{k \cdot q(x | \cntxt)}{n \cdot \exp{(-E(\bx)_t)} + k \cdot q(x | \cntxt)} =\\ &\sigma\left(E(\bx)_t + \log\left( \frac{k \cdot q(x | \cntxt)}{n} \right)\right)
}
While ELECTRA learns whether a token is more likely to come from the data distribution $\pdata$ or  noise distribution $q$, Electric only learns $\pdata$ because $q$ is passed into the model directly. This difference is analogous to using negative sampling \cite{Mikolov2013DistributedRO} vs. noise-contrastive estimation \cite{Mnih2013LearningWE} for learning word embeddings.

A disadvantage of Electric compared to ELECTRA is that it is less flexible in the choice of noise distribution. Since ELECTRA's binary classifier does not need to access $q$, its $q$ only needs to be defined for negative sample positions in the input sequence. Therefore ELECTRA can use a masked language model rather than a two-tower cloze model for $q$.
An advantage of Electric is that it directly provides (un-normalized) probabilities $\up$ for tokens, making it useful for applications such as re-ranking the outputs of text generation systems.
The differences between ELECTRA and Electric are summarized below: \\

\vspace{-3mm}

\small
\renewcommand{\arraystretch}{1.0}
\addtolength{\tabcolsep}{-3pt}
\noindent\begin{tabularx}{\linewidth}{X c c}
  \ttop
  Model & Noise Dist. & Binary Classifier  \tstrut \bstrut \\ \thinline
  Electric & \makecell{Two-Tower\\ Cloze Model} & $\sigma\left(E(\bx)_t + \log\left( \frac{k \cdot q(x | \cntxt)}{n} \right)\right)$ \tstrut  \\
  ELECTRA  & Masked LM \tstrut & $\sigma(E(\bx)_t)$ \tbottom
\end{tabularx} 
\addtolength{\tabcolsep}{3pt}
\renewcommand{\arraystretch}{1.0}
 \normalsize
\vspace{-1mm}

\section{Experiments}
We train two Electric models the same size as BERT-Base (110M parameters):
one on Wikipedia and BooksCorpus \cite{Zhu2015AligningBA} for comparison with BERT 
and one on OpenWebTextCorpus \cite{Gokaslan2019OpenWeb} for comparison\footnote{The original GPT-2 dataset is not public, so we use a public re-implementation.} with GPT-2.
The noise distribution transformers $T_{\textsc{ltr}}$ and $T_{\textsc{rtl}}$ are 1/4 the hidden size of Electric.
We do no hyperparameter tuning, using the same hyperparameter values as ELECTRA. 
Further details on training are in the appendix.

\addtolength{\tabcolsep}{-2pt}
\begin{table}[tb]
\small
\begin{center}
\begin{tabularx}{\linewidth}{X l l l l }
\ttop
\textbf{Model} & \textbf{MultiNLI} & \textbf{SQuAD 2.0} & \textbf{GLUE Avg.}  \tstrut \tsep
BERT & 84.3 & 73.7 & 82.2 \tstrut \\
XLNet & 85.8 & 78.5 & --  \\
ELECTRA & 86.2 & 80.5 & 85.1  \tsep 
Electric & 85.7 & 80.1 & 84.1  \tstrut 
\tbottom
\end{tabularx} 
\end{center}
\vspace{-0.5mm}
\caption{Dev-set scores of pre-trained models on downstream tasks. To provide direct comparisons, we only show base-sized models pre-trained on WikiBooks.}
\vspace{-1.5mm}
\label{tab:downstream}
\end{table}
\addtolength{\tabcolsep}{2pt}

\subsection{Transfer to Downstream Tasks}

We evaluate fine-tuning the Electric model on the GLUE natural language understanding benchmark \cite{wang2018glue} and the SQuAD 2.0 question answering dataset \cite{rajpurkar2018know}. 
We report exact-match for SQuAD, average score\footnote{Matthews correlation coefficient for CoLA, Spearman correlation for STS, accuracy for the other tasks.} over the GLUE tasks\footnote{We exclude WNLI, for which models do not outperform the majority baseline.}, and accuracy on the multi-genre natural language inference GLUE task.
Reported scores are medians over 10 fine-tuning runs with different random seeds.
We use the same fine-tuning setup and hyperparameters as ELECTRA.

Results are shown in Table~\ref{tab:downstream}.
Electric scores better than BERT, showing the energy-based formulation improves cloze model pre-training.
However, Electric scores slightly lower than ELECTRA.
One possible explanation is that Electric's noise distribution is worse because a two-tower cloze model is less expressive than a masked LM.
We tested this hypothesis by training ELECTRA with the same two-tower noise model as Electric. Performance did indeed go down, but it only explained about half the gap.
The surprising drop in performance suggests that learning the difference between the data and generations from a low-capacity model leads to better representations than only learning the data distribution, but we believe further research is needed to fully understand the discrepancy.   

\subsection{Fast Pseudo-Log-Likelihood Scoring}
\label{sec:pll}

An advantage of Electric over BERT is that it can efficiently produce pseudo-log-likelihood (PLL) scores for text \cite{Wang2019BERTHA}. PLLs for Electric are
\vspace{-1.5mm}
\alns{
    \text{PLL}(\bx) 
    \approx \sum_{t=1}^n \log(\up(x_t | \cntxt)) 
    = \sum_{t=1}^n -E(\bx)_t \vspace{-1mm}
}
\vspace{-0.5mm}
PLLs can be used to re-rank the outputs of an NMT or ASR system.
While historically log-likelihoods from language models have been used for such re-ranking, recent work has demonstrated that PLLs from masked language models perform better \cite{shin2019effective}. 
However, computing PLLs from a masked language model requires $n$ passes of the transformer: once with each token masked out.
\citet{Salazar2019MaskedLM} suggest distilling BERT into a model that uses no masking to avoid this cost, but this model considerably under-performed regular LMs in their experiments.

Electric can produce PLLs for all input tokens in a single pass like a LM while being bidirectional like a masked LM.
We use the PLLs from Electric for re-ranking the 100-best hypotheses of a 5-layer BLSTMP model 
from ESPnet \citep{Watanabe2018ESPnetES} on the 960-hour LibriSpeech corpus \citep{Panayotov2015LibrispeechAA} following the same experimental setup and using the same n-best lists as \citet{Salazar2019MaskedLM}.
Given speech features $\bs$ and speech recognition model $f$ the re-ranked output is
\vspace{-0.1mm}
\alns{
   \argmax_{\bx \in \text{n-best}(f, \bs)}  f(\bx|\bs) + \lambda \text{PLL}(\bx)
}
\vspace{-0.1mm}
where $\text{n-best}(f, \bs)$ consists of the top $n$ (we use $n=100$) predictions from the speech recognition model found with beam search, $f(\bx|\bs)$ is the score the speech model assigns the candidate output sequence $\bx$.
We select the best $\lambda$ on the dev set out of $[0.05, 0.1, ..., 0.95, 1.0]$, with different $\lambda$s selected for the ``clean" and ``other" portions of the data.

We compare Electric against GPT-2 \citep{Radford2019LanguageMA}, BERT \citep{devlin2018bert}, and two baseline systems that are bidirectional while only requiring a single transformer pass like Electric. TwoTower is a two-tower cloze model similar to Electric's noise distribution, but of the same size as Electric.
ELECTRA-TT is identical to ELECTRA except it uses a two-tower noise distribution rather than a masked language model.\footnote{With ELECTRA's original masked LM generator, it would be impossible to score all tokens in a single pass.} 
The noise distribution probabilities and binary classifiers scores of ELECTRA can be combined to assign probabilities for tokens as shown in Appendix G of the ELECTRA paper.

Results are shown in Table~\ref{tab:pseudo}.
Electric scores better than GPT-2 when trained on comparable data. 
While scoring worse than BERT, Electric is much faster to run.
It also slightly outperforms ELECTRA-TT, which is consistent with the finding from \citet{Labeau2018LearningWN} that NCE outperforms negative sampling for training language models.
Furthermore, Electric is simpler and faster than ELETRA-TT in that it does not require running the generator to produce PLL scores.
TwoTower scores lower than Electric, presumably because it is not a ``deeply" bidirectional model and instead just concatenates forward and backward hidden states.

\renewcommand{\arraystretch}{1.1}
\addtolength{\tabcolsep}{-2.2pt}
\begin{table}[tb]
\small
\begin{center}
\begin{tabularx}{1.0\linewidth}{X l l l c}
\ttop 
\multirow{2}{*}{\textbf{Model}} & \textbf{Pre-train} & \textbf{Clean} & \textbf{Other} & \multirow{2}{*}{\textbf{Runtime}} \tstrut \\
 & \textbf{Data} & \textbf{WER} & \textbf{WER} &  \tsep
  None & -- & 7.26 & 20.37  & 0\tstrut \tsep
  BERT  & WikiBooks & 5.41 & 17.41 & $n$\tstrut\\
  Electric  & WikiBooks & 5.65 & 17.42 & 1\tsep
  GPT-2  & OWT & 5.64 & 17.60  & 1\tstrut\\
    TwoTower & OWT*  & 5.32 & 17.25 & 1\\
   ELECTRA-TT & OWT*  & 5.22 & 17.01 & 1\\
  Electric & OWT* & 5.18 & 16.93 &  1\tbottom
\end{tabularx} 
\end{center}
\vspace{-0.5mm}
\caption{
Test-set word error rates on LibriSpeech after rescoring with base-sized models. None, GPT-2, and BERT results are from \citet{Salazar2019MaskedLM}.
Runtime is measured in passes through the transformer.
``Clean" and ``other" are easier and harder splits of the data. *We use a public re-implementation of OpenWebText. 
}
\vspace{-0.9mm}
\label{tab:pseudo}
\end{table}
\addtolength{\tabcolsep}{2.2pt}
\renewcommand{\arraystretch}{1.0}

\section{Related Work}

Language modeling \cite{dai2015semi,radford2018improving,peters2018deep} and cloze modeling \cite{devlin2018bert,baevski2019cloze,liu2019roberta} have proven to be effective pre-training tasks for NLP.
Unlike Electric, these methods follow the standard recipe of estimating token probabilities with an output softmax and using maximum-likelihood training.

Energy-based models have been widely
explored in machine learning \citep{Dayan1995TheHM,LeCun2006ATO}.
While many training methods involve sampling from the EBM using gradient-based MCMC \citep{Du2019ImplicitGA} or Gibbs sampling \citep{Hinton2002TrainingPO}, we considered these approaches too slow for pre-training because they require multiple passes through the model per sample.
We instead use noise-contrastive estimation \citep{Gutmann2010NoisecontrastiveEA}, which has widely been used in NLP for learning word vectors \cite{Mnih2013LearningWE} and text generation models \cite{Jean2014OnUV, Jzefowicz2016ExploringTL}.
While EBMs have previously been applied to left-to-right \citep{Wang2015TransdimensionalRF} or globally normalized \citep{Rosenfeld2001WholesentenceEL, Deng2020ResidualEM} text generation, they have not previously been applied to cloze models or for pre-training NLP models. 
Several papers have pointed out the connection between EBMs and GANs \citep{zhao2016energy,finn2016connection}, which is similar to the Electric/ELECTRA connection.

\section{Conclusion}
We have developed an energy-based cloze model we call Electric and designed an efficient training algorithm for Electric based on noise-contrastive estimation.
Although Electric can be derived solely from the cloze task, the resulting pre-training method is closely related to ELECTRA's GAN-like pre-training algorithm.
While slightly under-performing ELECTRA on downstream tasks, Electric is useful for its ability to quickly produce pseudo-log-likelihood scores for text. 
Furthermore, it offers a clearer and more principled view of the ELECTRA objective as a ``negative sampling" version of cloze pre-training. 

\section*{Acknowledgements}
We thank John Hewitt, Yuhao Zhang, Ashwin Paranjape, Sergey Levine, and the  anonymous reviewers for their thoughtful comments and suggestions.
Kevin is supported by a Google PhD Fellowship.

\bibliographystyle{acl_natbib}
\bibliography{emnlp2020.bib}

\appendix

\section{Pre-Training Details}

The neural architectures of our models are identical to BERT-Base \citep{devlin2018bert}, although we believe incorporating additions such as relative position encodings \citep{Shaw2018SelfAttentionWR} would improve results. Our pre-training setup is the same as ELECTRA's \citep{clark2020electra}, which adds some additional ideas from \citet{liu2019roberta} on top of the BERT codebase, such as dynamic masking and removing the next-sentence prediction task.
We use the weight sharing trick from ELECTRA, where the transformers producing the proposal distribution and the main transformer share token embeddings. 
We do not use whole-word or n-gram masking, although we believe it would improve results too. 

We did no hyperparameter tuning, directly using the hyperparameters from ELECTRA-Base for Electric and our baselines.
These hyperparameters are slightly modified from the ones used in BERT; for completeness, we show these values in Table~\ref{tab:hparams}.
The hidden sizes, feed-forward hidden sizes, and number of attention heads of the two transformers $T_{\textsc{ltr}}$ and $T_{\textsc{rtl}}$ used to produce the proposal distribution are 1/4 the size of Electric. We chose this value because it keeps the compute comparable to ELECTRA; running two 1/4-sized transformers takes roughly the same compute as running one 1/3-sized transformer, which is the size of ELECTRA's generator. To make the compute exactly equal, we train Electric for slightly fewer steps than ELECTRA. 
This same generator architecture was used for ELECTRA-TT.
The TwoTower baseline consists of two transformers 2/3 the size of BERT's, which takes approximately the same compute to run. 
The Electric models, ELECTRA-Base, and BERT-Base all use the same amount of pre-train compute (e.g., Electric is trained for fewer steps than BERT due to the extra compute from the proposal distribution), which equates to approximately three days of training on 16 TPUv2s.

\section{Fine-Tuning Details}

We use ELECTRA's top-level classifiers and hyperparameter values for fine-tuning as well.
For GLUE tasks, a simple linear classifier is added on top of the pre-trained transformer. For SQuAD, a question answering module similar XLNet's \citep{yang2019xlnet} is added on top of the transformer, which is slightly more sophisticated than
BERT’s in that it jointly rather than independently predicts the start and end positions and has an “answerability” classifier added for SQuAD 2.0. ELECTRA's hyperparameters are similar to BERT's, with the main difference being the addition of a layer-wise learning rate decay where layers of the network closer to the output have a higher learning rate. 
Following BERT, we submit the best of 10 models fine-tuned with different random seeds to the GLUE leaderboard for test set results.

\addtolength{\tabcolsep}{0pt}
\begin{table*}[t!]
\small
\begin{center}
\begin{tabularx}{0.7\linewidth}{X l l}
\ttop
\textbf{Hyperparameter} & \textbf{Pre-Training} & \textbf{Fine-Tuning} \tstrut \tsep
Number of layers & \multicolumn{2}{l}{12}   \tstrut \\
Hidden Size & \multicolumn{2}{l}{768}  \\
FFN inner hidden size & \multicolumn{2}{l}{3072}  \\
Attention heads & \multicolumn{2}{l}{12}  \\
Attention head size & \multicolumn{2}{l}{64} \\
Embedding Size & \multicolumn{2}{l}{768}  \\
Proposal Transformer Size & 1/4 & NA  \\
Negative sample percent & 15 & NA  \\
Learning Rate Decay & \multicolumn{2}{l}{Linear} \\
Warmup steps & 10000 & First 10\%  \\
Learning Rate & 5e-4 & 1e-4  \\
Layerwise LR decay & None & 0.8  \\
 Adam $\epsilon$ & \multicolumn{2}{l}{1e-6}  \\
 Adam $\beta_1$ & \multicolumn{2}{l}{0.9}  \\
 Adam $\beta_2$ & \multicolumn{2}{l}{0.999} \\
 Attention Dropout & \multicolumn{2}{l}{0.1}  \\
 Dropout & \multicolumn{2}{l}{0.1}  \\
 Weight Decay & 0.01 & 0  \\
 Batch Size & 256 & 32 \\
 \multirow{2}{*}{Train Steps}  & \multirow{2}{*}{700K} & 10 epochs for RTE and STS \\
  & & 2 for SQuAD, 3 for other tasks  \tbottom
\end{tabularx} 
\end{center}
\caption{Hyperparameters for Electric; the values are identical to ELECTRA's other than the train steps and different-sized proposal network (see text), but we include them here for completeness. If not shown, the fine-tuning hyperparameter is the same as the pre-training one.}
\label{tab:hparams}
\end{table*}
\addtolength{\tabcolsep}{0pt}

\section{Dataset Details}
We provide details on the fine-tuning datasets below. 
All datasets are in English. 
GLUE data can be downloaded at \url{https://gluebenchmark.com/} and SQuAD data can be downloaded at \url{https://rajpurkar.github.io/SQuAD-explorer/}.
\begin{itemize}
    \item \textbf{CoLA:} Corpus of Linguistic Acceptability \citep{Warstadt2018NeuralNA}. The task is to determine whether a given sentence is grammatical or not. The dataset contains 8.5k train examples from books and journal articles on linguistic theory. 
    \item \textbf{SST:} Stanford Sentiment Treebank \citep{Socher2013RecursiveDM}. The tasks is to determine if the sentence is positive or negative in sentiment. The dataset contains 67k train examples from movie reviews.
    \item \textbf{MRPC:} Microsoft Research Paraphrase Corpus \citep{Dolan2005AutomaticallyCA}. The task is to predict whether two sentences are semantically equivalent or not. The dataset contains 3.7k train examples from online news sources.
    \item \textbf{STS:} Semantic Textual Similarity \citep{Cer2017SemEval2017T1}. The tasks is to predict how semantically similar two sentences are on a 1-5 scale. The dataset contains 5.8k train examples drawn from new headlines, video and image captions, and natural language inference data.
    \item \textbf{QQP:} Quora Question Pairs \citep{QQP}. The task is to determine whether a pair of questions are semantically equivalent. The dataset contains 364k train examples from the community question-answering website Quora.
    \item \textbf{MNLI:} Multi-genre Natural Language Inference \citep{Williams2018ABC}. Given a premise sentence and a hypothesis sentence, the task is to predict whether the premise entails the hypothesis, contradicts the hypothesis, or neither. The dataset contains 393k train examples drawn from ten different sources.
    \item \textbf{QNLI:} Question Natural Language Inference; constructed from SQuAD \citep{Rajpurkar2016SQuAD10}. The task is to predict whether a context sentence contains the answer to a question sentence. The dataset contains 108k train examples from Wikipedia.
    \item \textbf{RTE:} Recognizing Textual Entailment \citep{Giampiccolo2007TheTP}. Given a premise sentence and a hypothesis sentence, the task is to predict whether the premise entails the hypothesis or not. The dataset contains 2.5k train examples from a series of annual textual entailment challenges.
    \item \textbf{SQuAD 1.1:} Stanford Question Answering Dataset \citep{Rajpurkar2016SQuAD10}. Given a context paragraph and a question, the task is to select the span of text in the paragraph answering the question. The dataset contains 88k train examples from Wikipedia.
    \item \textbf{SQuAD 2.0:} Stanford Question Answering Dataset version 2.0 \citep{rajpurkar2018know}. This task adds addition questions to SQuAD whose answer does not exist in the context; models have to recognize when these questions occur and not return an answer for them. The dataset contains 130k train examples from Wikipedia.
\end{itemize}

We report Spearman correlation for STS, Matthews correlation coefficient (MCC) for CoLA, exact match for SQuAD, and accuracy for the other tasks. 
We use the provided evaluation script for SQuAD\footnote{\url{https://worksheets.codalab.org/rest/bundles/0x6b567e1cf2e041ec80d7098f031c5c9e/contents/blob/}}, scipy to compute Spearman scores\footnote{\url{https://docs.scipy.org/doc/scipy/reference/generated/scipy.stats.spearmanr.html}}, and sklearn to compute MCC\footnote{\url{https://scikit-learn.org/stable/modules/generated/sklearn.metrics.matthews_corrcoef.html}}. 
We use the standard train/dev/test splits.

\addtolength{\tabcolsep}{-2pt}
\begin{table*}[tb]
\small
\begin{center}
\begin{tabularx}{\linewidth}{X l l l l l l l l l l}
\ttop 
\textbf{Model} & \textbf{CoLA} & \textbf{SST} & \textbf{MRPC} & \textbf{STS} & \textbf{QQP} & \textbf{MNLI} & \textbf{QNLI} & \textbf{RTE} & \textbf{SQuAD 1.1} &  \textbf{SQuAD 2.0} \tstrut \tsep 
& MCC & Acc & Acc & Spear & Acc & Acc & Acc & Acc & EM &  EM \tstrut \tsep 
\multicolumn{11}{l}{\it{Dev set results}} \tstrut \\
BERT & 58.4 & 92.8 & 86.0 & 87.8 & 90.8 & 84.5 & 88.6 & 68.5 & 80.8 & 73.7  \\
XLNet & -- & 93.4 & -- & -- & -- & 85.8 & -- & -- & -- & 78.5 \\
ELECTRA & 65.8 & 92.4 & 87.9 & 89.1 & 90.9 & 86.2 & 92.4 & 76.3 & 84.5 & 80.5 \tsep
Electric & 61.8 & 91.9 & 88.0 & 89.4 & 90.6 & 85.7 & 92.1 & 73.4 & 84.5 & 80.1 \tstrut \tbottom
\multicolumn{11}{l}{\it{Test set results}} \tstrut \\
BERT & 52.1 & 93.5 & 84.8 & 85.8 & 89.2 & 84.6 & 90.5 & 66.4 & --& -- \\
ELECTRA & 59.7 & 93.4 & 86.7 & 87.7 & 89.1 & 85.8 & 92.7 & 73.1 & -- & -- \tsep
Electric & 61.5 & 93.2 & 85.4 & 86.9 & 89.2 & 85.2 & 91.8 & 67.3 & -- & -- \tstrut \tbottom
\end{tabularx} 
\end{center}
\caption{GLUE scores pre-trained models on downstream tasks. To provide direct comparisons, we only show base-sized models pre-trained on WikiBooks and fine-tuned with standard single-task training.}
\label{tab:gluefull}
\end{table*}
\addtolength{\tabcolsep}{2pt}

\addtolength{\tabcolsep}{3pt}
\begin{table*}[t!]
\begin{center}
\begin{tabularx}{0.96\linewidth}{X l l l l l c}
\ttop 
\multirow{2}{*}{\textbf{Rescoring Model}} & \textbf{Pre-Training} & \multicolumn{2}{c}{\textbf{Dev}} & \multicolumn{2}{c}{\textbf{Test}} & \textbf{Transformer} \tstrut \\
  & \textbf{Data} & clean & other & clean & other  &  \textbf{Passes} \tsep
  None & -- & 7.17 & 19.79 & 7.26 & 20.37 & 0\tstrut\tsep
  BERT & WikiBooks & 5.17 & 16.44 & 5.41 & 17.41 & $n$\tstrut\\
    Electric & Wikibooks & 5.47 & 16.56 & 5.65 & 17.42 & 1\tsep
  GPT-2 & OpenWebText & 5.39 & 16.81 & 5.64 & 17.61 & 1\tstrut\\
  TwoTower & OpenWebText & 5.12 & 16.37 & 5.32 & 17.25 & 1\\
  ELECTRA-TT & OpenWebText & 5.05 & 16.27 & 5.22 & 17.01 & 1\\
  Electric & OpenWebText & 4.97 & 16.23 & 5.18 & 16.93 & 1\tbottom
\end{tabularx} 
\end{center}
\caption{Word error rates on LibriSpeech after rescoring with base-sized models. None, GPT-2, and BERT results are from \citet{Salazar2019MaskedLM}.
Runtime is measured in passes through the transformer and data indicates the pre-training dataset.
``Clean" and ``other" are easier and harder splits of the data. *We use a public re-implementation of OpenWebText.}
\label{tab:asrfull}
\end{table*}
\addtolength{\tabcolsep}{-3pt}

\section{Detailed Results}
We show detailed results on GLUE and SQuAD in Table~\ref{tab:gluefull} and detailed results on LibriSpeech re-ranking in Table~\ref{tab:asrfull}.
Following BERT, we do not show results on the WNLI GLUE task, as it is
difficult to beat even the majority classifier using a standard fine-tuning-as-classifier approach.
We show dev rather than test results on GLUE in the main paper because they are more reliable; the performance of fine-tuned models varies substantially based on the random seed \citep{phang2018sentence,Clark2019BAMBM,dodge2020fine}, but GLUE only supports submitting a single model rather than getting a median score of multiple models.
While using dev-set model selection to choose the test set submission may alleviate the high variance of fine-tuning to some extent, such model selection is still not sufficient for reliable comparisons between methods \citep{reimers2018comparing}.

\end{document}